\documentclass[conference]{IEEEtran}
\IEEEoverridecommandlockouts
\usepackage{cite}
\usepackage{amsmath,amssymb,amsfonts}
\usepackage{algorithmic}
\usepackage{graphicx}
\usepackage{textcomp}
\usepackage{xcolor}
\usepackage{booktabs}
\usepackage{threeparttable}
\usepackage[normalem]{ulem}
\usepackage{subcaption}
\usepackage{makecell}
\usepackage{balance}
\usepackage{url}
\usepackage{hyperref}
\hypersetup{
    colorlinks=true,
    urlcolor=blue
}

\def\BibTeX{{\rm B\kern-.05em{\sc i\kern-.025em b}\kern-.08em
    T\kern-.1667em\lower.7ex\hbox{E}\kern-.125emX}}

\usepackage{multirow}
\usepackage{pifont}
\usepackage{float}

\begin{document}

\bstctlcite{IEEEexample:BSTcontrol}
\title{BAP-MOS: Bandit-Based Adaptive Prompting for Boundary-Sensitive Multi-Organ Segmentation
}

\author{
\IEEEauthorblockN{
Satvik Praveen\textsuperscript{1},
Shengji Jin\textsuperscript{1},
Ahmed Lamidi\textsuperscript{1},\\
Xin Qian\textsuperscript{2},
Yi Sheng\textsuperscript{1}
}

\IEEEauthorblockA{
\textsuperscript{1}\textit{Computer Science \& Engineering, University of South Florida, Tampa, USA}\\
\textsuperscript{2}\textit{Department of Physics and Astronomy, Stony Brook University, Stony Brook, USA}\\
\{satvikpraveen,jins,ahmedlamidi\}@usf.edu,
Xin.Qian@stonybrookmedicine.edu,
sheng1@usf.edu
}
}

\maketitle

\begingroup
\renewcommand{\thefootnote}{}
\footnotetext{\footnotesize
This work has been submitted to the IEEE for possible publication.
Copyright may be transferred without notice, after which this version
may no longer be accessible.}
\addtocounter{footnote}{-1}
\endgroup

\begin{abstract}

Multi-organ ultrasound segmentation remains challenging when anatomically adjacent structures must be delineated jointly, as localized boundary errors can persist even when Dice scores are high. To address these challenges, we propose Boundary-Adaptive Prompting for Multi-Organ Segmentation (BAP-MOS), a closed-loop adaptive prompting framework. BAP-MOS formulates prompt selection as an organ-specific multi-armed bandit problem over box, point, and combined prompts. An outer Tree-structured Parzen Estimator (TPE) loop selects the prompt-selection parameter vector, while an inner UCB-Tuned loop adapts per-organ prompt preferences during fine-tuning using a bounded Dice--MSD--HD95 validation-probe reward. The framework further introduces an organ-scaled negative prompt ring to adapt sparse prompt geometry across anatomical scales, while keeping the image and prompt encoders frozen and updating only the mask decoder. We evaluate BAP-MOS on pooled prostate-region TRUS cohorts against U-Net, nnU-Net, MedSAM, fixed-prompt SAM/MedSAM, and adaptive policy variants. On this benchmark, BAP-MOS achieves Dice \textit{0.982}, HD95 \textit{0.482}, and MSD \textit{0.204}, reducing HD95 by approximately \textit{48\%} and MSD by \textit{45\%} relative to the strongest conventional baseline. To verify the generalization ability of the framework, we tested it on the external PFUS1 pelvic-floor ultrasound corpus using MedSAM and its adaptive strategy variants, and the results were good. These results support adaptive prompt allocation as an effective mechanism for improving boundary-sensitive multi-organ ultrasound segmentation without modifying the foundation-model backbone.
Source Code is available at: \url{https://github.com/SatvikPraveen/BAP-MOS}

\color{black}

\end{abstract}

\begin{IEEEkeywords}
Multi-organ segmentation, adaptive prompting, multi-armed bandits, boundary-aware optimization.
\end{IEEEkeywords}

\section{Introduction}

Ultrasound is one of the most widely used clinical imaging modalities because it is inexpensive, portable, real-time, and free of ionizing radiation~\cite{noble2006ultrasound}, making it indispensable for image-guided procedures such as prostate biopsy, brachytherapy, pelvic-floor assessment, and transperineal interventions~\cite{lei2021male, yang2021deep, solis2025pfus1}. Unlike single-structure tasks, these procedures require interpreting multiple adjacent organs, such as the prostate, bladder, rectum, and urethra, to localize treatment targets while protecting neighboring organs at risk. Ultrasound segmentation has therefore evolved from single- to multi-organ settings, which pose fundamentally different challenges.

\begin{figure}[t]
  \centering
  \includegraphics[width=0.4\textwidth]{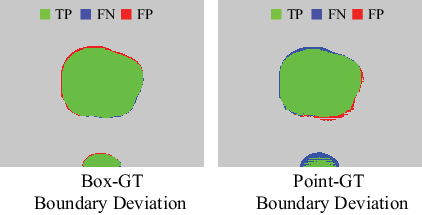}
  \caption{High Dice may hide boundary errors. Box and Point prompts achieve similar region overlap but exhibit substantially different boundary deviations.}
  \label{fig:boundary_problem}
  \vspace{-0.5cm}
\end{figure}

Despite the rapid progress of deep learning, accurately segmenting multiple adjacent organs in ultrasound remains difficult. Existing methods often report excellent Dice scores, suggesting that organ regions are successfully identified. However, this metric largely reflects region overlap and provides limited information about the quality of shared inter-organ boundaries~\cite{taha2015metrics, yeghiazaryan2018boundary_metrics}. As illustrated in Fig.~\ref{fig:boundary_problem}, two segmentation results with similarly plausible organ masks may exhibit substantially different boundary behavior. While the organ interiors overlap well with the ground truth, contour deviations accumulate precisely along the interfaces between neighboring organs, resulting in over-segmentation and under-segmentation that are largely invisible to Dice. This phenomenon is clinically important because image-guided interventions depend primarily on accurate boundary localization rather than overall region overlap. In prostate brachytherapy, for example, small boundary deviations between the prostate and rectum may lead to inaccurate localization of the organ interface, potentially affecting downstream image-guided procedures and quantitative measurements~\cite{henry2022gec}.

These observations suggest that the primary challenge of multi-organ ultrasound segmentation lies not only in recognizing individual organs, but also in accurately delineating the shared interfaces between adjacent structures. Prior methods mainly address this through stronger architectures~\cite{ronneberger2015u, milletari2016v, isensee2021nnu} or boundary-aware losses~\cite{kervadec2019boundary, karimi2019reducing}, but leave untouched an additional degree of freedom offered by promptable foundation models~\cite{Kirillov_2023_ICCV}: how the model is guided during segmentation, where the same backbone yields different predictions under different prompts~\cite{mazurowski2023segment, he2023computer}. Our empirical analysis shows that boundary quality is highly sensitive to this guidance, suggesting that prompting should be an adaptive component of segmentation rather than a fixed design choice.

Motivated by these observations, we propose Boundary-Adaptive Prompting for Multi-Organ Segmentation (BAP-MOS), a lightweight boundary-aware adaptation framework for promptable segmentation models. Instead of costly full-model fine-tuning, BAP-MOS formulates prompt selection as a per-organ sequential decision process~\cite{auer2002finite} guided by boundary-sensitive feedback, built on a frozen encoder with lightweight decoder adaptation. It dynamically selects the prompting strategy that best preserves organ boundaries while remaining compatible with different promptable backbones, including SAM~\cite{Kirillov_2023_ICCV} and MedSAM~\cite{ma2024segment}. By optimizing boundary fidelity rather than only region overlap, BAP-MOS improves clinically relevant segmentation quality without modifying the underlying model. Our main contributions are as follows:

\begin{itemize}
    \item \textbf{Boundary-oriented empirical analysis.} We identify a failure mode in multi-organ ultrasound segmentation, showing that boundary degradation can remain severe despite consistently high Dice scores, and demonstrate that boundary quality is substantially more sensitive to prompting strategy than conventional overlap metrics suggest.
    \item \textbf{Boundary-Adaptive Prompting for Multi-Organ Segmentation.} We propose BAP-MOS, a lightweight framework that formulates prompt selection as a per-organ adaptive decision driven by boundary-sensitive feedback while retaining an efficient frozen-backbone architecture.
    \item \textbf{Comprehensive validation.} We evaluate BAP-MOS on both prostate TRUS and pelvic-floor ultrasound datasets using SAM and MedSAM backbones, demonstrating consistent improvements in boundary accuracy over conventional segmentation networks and fixed prompting strategies while maintaining competitive Dice performance.
\end{itemize}

\color{black}

\section{Related Work}


\subsection{Multi-Organ Ultrasound Segmentation}

Multi-organ ultrasound segmentation has been approached predominantly with deep encoder-decoder networks~\cite{liu2024towards}, from U-Net~\cite{ronneberger2015u} and its variants, including V-Net~\cite{milletari2016v} and attention-based designs~\cite{li2025hfa}, to self-configuring pipelines such as nnU-Net~\cite{isensee2021nnu}, in settings that include prostate TRUS and pelvic-floor imaging~\cite{lei2021male, yang2021deep, solis2025pfus1, jiang2024microsegnet, jiao2025u, szentimrey2023automated}. These models differ in architecture and supervision, but share a static mapping from image appearance to per-pixel labels, and are commonly optimized and selected primarily around regional agreement. Ultrasound makes the output especially fragile at the boundary, because speckle, acoustic shadowing, and weak tissue contrast blur the very interfaces that separate adjacent organs \cite{noble2006ultrasound}. These artifacts make ultrasound harder than higher-contrast modalities, but they are not why boundaries fail here.

The failure is methodological: these methods optimize for regional overlap, and the inter-organ boundary is never something they are trained to control. This begins with measurement. Dice~\cite{dice1945measures} and related overlap scores aggregate agreement over an entire region and are by construction insensitive to localized contour error, particularly when a deviation occupies only a thin band along a large structure; this is the established reason the field turns to surface-distance metrics such as mean surface distance (MSD)~\cite{heimann2009comparison} and the 95th-percentile Hausdorff distance (HD95)~\cite{huttenlocher2002comparing} to quantify boundary fidelity \cite{taha2015metrics, yeghiazaryan2018boundary_metrics}. In the multi-organ case it is most severe, because the boundaries of greatest clinical importance are the low-contrast interfaces shared between organs, precisely the errors an overlap objective is least able to see.

What the literature therefore lacks is not another overlap-optimized model, but a way to treat the inter-organ boundary as an explicit target of adaptation rather than a byproduct of overlap. That requires a control point at which each organ's delineation can be steered in response to boundary quality, rather than fixed once the model is trained.

\vspace{-5pt}
\subsection{Promptable Foundation Models}

Promptable foundation models introduce such a steering interface. The Segment Anything Model (SAM)~\cite{Kirillov_2023_ICCV} conditions mask prediction on spatial prompts such as points and bounding boxes rather than on the image alone, making the prompt an additional degree of freedom that steers a fixed backbone toward different masks. Medical adaptations and evaluations of SAM, including MedSAM~\cite{ma2024segment} and related studies, show both the promise of this interface and its sensitivity to imaging modality, object appearance, prompt type, and prompt placement~\cite{mazurowski2023segment, he2023computer}.

Recent work has therefore focused on how prompts should be obtained. Some approaches derive prompts from weak supervision, such as image-level labels~\cite{wang2025weakmedsam, zou2025acea}. Others learn auxiliary prompt generators that predict box, mask, or embedding prompts from the image, including in medical ultrasound~\cite{wahd2025sam2rad}, or automatically generate the prompts needed for multi-organ segmentation without any manual input~\cite{li2025autoprosam}. These studies move beyond architecture changes by exploiting the prompt as a controllable input to a foundation model.

Two limitations persist, however. First, like the region-based methods above, this work is still evaluated and driven by overlap, so the prompt is not optimized with respect to boundary distances such as MSD or HD95. Second, the effort concentrated on the prompt goes into producing one, deriving or predicting it, rather than choosing among the point, box, and combined prompts already available; these methods optimize how a prompt is generated, leaving how to select and adapt one to be taken up next.

\subsection{Adaptive Prompting}

Adaptive prompting lets a policy adjust the prompt in response to the model's own feedback. One line adjusts where the prompt is placed: a reinforcement-learning agent can iteratively supply informative prompts to a frozen SAM so that its predictions align with a target task \cite{huang2024alignsam}, and a policy trained with proximal optimization can reposition point prompts to sharpen the resulting mask \cite{liu2025plug}. A second line adjusts which form of prompt is used: cast as a Markov decision process, an agent can select among prompt forms such as points and boxes over a sequence of interactions \cite{shen2023temporally}, and a later variant additionally learns when to stop, trading segmentation accuracy against interaction cost \cite{huang2024optimizing}. Together, these methods show that both the placement and the form of the prompt can be chosen adaptively.

Two properties are nonetheless shared across this line. The adaptation is global: a single policy is applied to whatever is being segmented, without differentiating the choice by anatomical structure, even though the organs in one multi-organ ultrasound image differ sharply in size, contrast, and boundary ambiguity, so that the form or placement best suited to a large high-contrast structure need not suit a small, weakly visualized one. And the objective is region accuracy, task alignment, or interaction efficiency rather than boundary quality, so the prompt is never adjusted toward better surface agreement at inter-organ boundaries.

Across all three groups the same gap remains: multi-organ segmentation optimizes overlap and leaves the boundary a byproduct; promptable models add the prompt as a control interface but only generate it; and adaptive prompting adjusts the prompt globally, by accuracy or efficiency rather than boundary quality, and never per organ. What is missing is a prompt decision that is at once organ-specific and driven by boundary-distance feedback as the model adapts, which is the gap BAP-MOS sets out to close.

\color{black}

\begin{table}[t]
\vspace{-0.5cm}
\centering
\scriptsize
\setlength{\tabcolsep}{5pt}
\caption{Decoder-only ($\sim$4M params) vs full-model ($\sim$93M params) fine-tuning across single and multiple Organ segmentation.}
\label{tab:decoder_vs_full}
\resizebox{\linewidth}{!}{
\begin{tabular}{ccccccc}
\toprule
\textbf{Type} & \textbf{Model} & \textbf{Prompt} & \textbf{Time(m)\,$\downarrow$} & \textbf{Dice\,$\uparrow$} & \textbf{HD95\,$\downarrow$} & \textbf{MSD\,$\downarrow$} \\
\midrule
\multirow{4}{*}{\begin{tabular}[c]{@{}c@{}}Single\\ Organ\end{tabular}} & \multirow{2}{*}{Decoder}    & Box    & 31 & 0.9937 & 2.84 & 1.09 \\
                                                                        &                             & Points & 36 & 0.9885 & 3.80 & 1.49 \\ \cmidrule(l){2-7}
                                                                        & \multirow{2}{*}{Full Model} & Box    & 63 & 0.9926 & 3.18 & 1.24 \\
                                                                        &                             & Points & 57 & 0.9852 & 4.52 & 1.68 \\
\midrule
\multirow{4}{*}{\begin{tabular}[c]{@{}c@{}}Multi\\ Organ\end{tabular}}  & \multirow{2}{*}{Decoder}    & Box    & 30 & 0.9735 & 4.21 & 1.77 \\
                                                                        &                             & Points & 53 & 0.9691 & 32.89 & 12.06 \\ \cmidrule(l){2-7}
                                                                        & \multirow{2}{*}{Full Model} & Box    & 64 & 0.9721 & 4.75 & 2.03 \\
                                                                        &                             & Points & 65 & 0.9722 & 5.34 & 2.15 \\
\bottomrule
\end{tabular}
}
\vspace{-0.5cm}
\end{table}




\color{black}

\section{Motivation}


Our design is motivated by two observations from multi-organ TRUS segmentation. First, full-model fine-tuning offers little advantage over lightweight decoder-only adaptation; additional trainable parameters therefore mainly increase cost rather than accuracy. Second, segmentation performance depends strongly on the prompting strategy, motivating a closer investigation of prompt selection.

\subsection{ Why Decoder-only}

We first investigate whether full-model fine-tuning is necessary for adapting SAM to multi-organ TRUS segmentation. Table~\ref{tab:decoder_vs_full} compares decoder-only and full-model fine-tuning under different prompting strategies for both single- and multi-organ segmentation.

In the \textbf{single-organ setting}, decoder-only and full-model fine-tuning perform comparably across Box and Point prompts. Full-model tuning nearly doubles the optimization time (63 vs. 31 minutes for Box) with little gain in Dice or boundary accuracy, indicating that decoder-only adaptation already offers an effective efficiency--accuracy trade-off for well-defined single-organ anatomy.


The \textbf{multi-organ setting} differs. Box prompts stay robust under both strategies, but Point prompts degrade severely under decoder-only adaptation despite a competitive Dice (0.9691). Full-model tuning alleviates this (HD95: 32.89 to 5.34 mm) at over 20× more trainable parameters and roughly 2× longer training. Increasing capacity improves robustness only at substantial cost; we therefore retain decoder-only adaptation as our backbone. More importantly, performance is highly sensitive to prompt choice, motivating the prompt-selection analysis next.

\begin{table}[t]
\vspace{-0.5cm}
\centering
\scriptsize
\begin{threeparttable}
\caption{Multi-organ TRUS segmentation under fixed prompt schedules.}
\label{tab:prompt_ratio}
\setlength{\tabcolsep}{5pt}
\begin{tabular}{cccc}
\toprule
\makecell{\textbf{Schedules}\\\textbf{(B:P:Bo)}} & \textbf{Dice\,$\uparrow$} & \textbf{HD95\,$\downarrow$} & \textbf{MSD\,$\downarrow$} \\
\midrule
1\,:\,9\,:\,0   & 0.956 \tiny{$\pm$0.016} & 1.026 \tiny{$\pm$0.188} & 0.419 \tiny{$\pm$0.085} \\
1\,:\,1\,:\,0   & 0.978 \tiny{$\pm$0.001} & 0.574 \tiny{$\pm$0.050} & 0.235 \tiny{$\pm$0.010} \\
9\,:\,1\,:\,0   & 0.980 \tiny{$\pm$0.001} & 0.531 \tiny{$\pm$0.040} & 0.223 \tiny{$\pm$0.013} \\
1\,:\,1\,:\,1   & 0.977 \tiny{$\pm$0.005} & 0.624 \tiny{$\pm$0.159} & 0.256 \tiny{$\pm$0.060} \\
1\,:\,5\,:\,1   & 0.958 \tiny{$\pm$0.011} & 0.982 \tiny{$\pm$0.150} & 0.416 \tiny{$\pm$0.074} \\
2\,:\,3\,:\,5   & \textbf{0.980} \tiny{$\pm$0.001} & \textbf{0.517} \tiny{$\pm$0.043} & \textbf{0.215} \tiny{$\pm$0.015} \\
5\,:\,1\,:\,1   & 0.978 \tiny{$\pm$0.001} & 0.601 \tiny{$\pm$0.044} & 0.243 \tiny{$\pm$0.016} \\
\midrule
\textbf{Spread} & \textbf{+2.5\%} & \textbf{+98\%} & \textbf{+95\%} \\
\bottomrule
\end{tabular}
\begin{tablenotes}
\footnotesize
\item[$\bullet$] Schedules are Box:Point:Both (B:P:Bo) ratios.
\item[$\bullet$] \textbf{Spread} $=(\max-\min)/\min$ across schedules.
\end{tablenotes}
\end{threeparttable}
\vspace{-0.3cm}
\end{table}

\subsection{ Why Prompt}

Having established decoder-only adaptation as an efficient backbone, we next investigate the effect of prompt scheduling. Table~\ref{tab:prompt_ratio} compares several fixed Box:Point:Both schedules and reveals a clear pattern. While Dice changes only marginally across different schedules (2.5\%), boundary metrics vary dramatically, with MSD and HD95 differing by 95\% and 98\%, respectively. In other words, prompt scheduling has little influence on region overlap but has a profound effect on clinically relevant boundary accuracy. This suggests that Dice alone is insufficient for selecting prompting strategies and motivates a boundary-aware view of prompt design.


\section{Method}
\label{sec:method}

BAP-MOS adapts a frozen promptable backbone to multi-organ ultrasound by deciding, per organ, which prompt best preserves its boundary, rather than by retraining the network or fixing a single prompt type.

\subsection{ TPE-Based Outer-Loop Optimization}

\begin{figure}[t]
\vspace{-0.5cm}
    \includegraphics[width=0.49\textwidth]{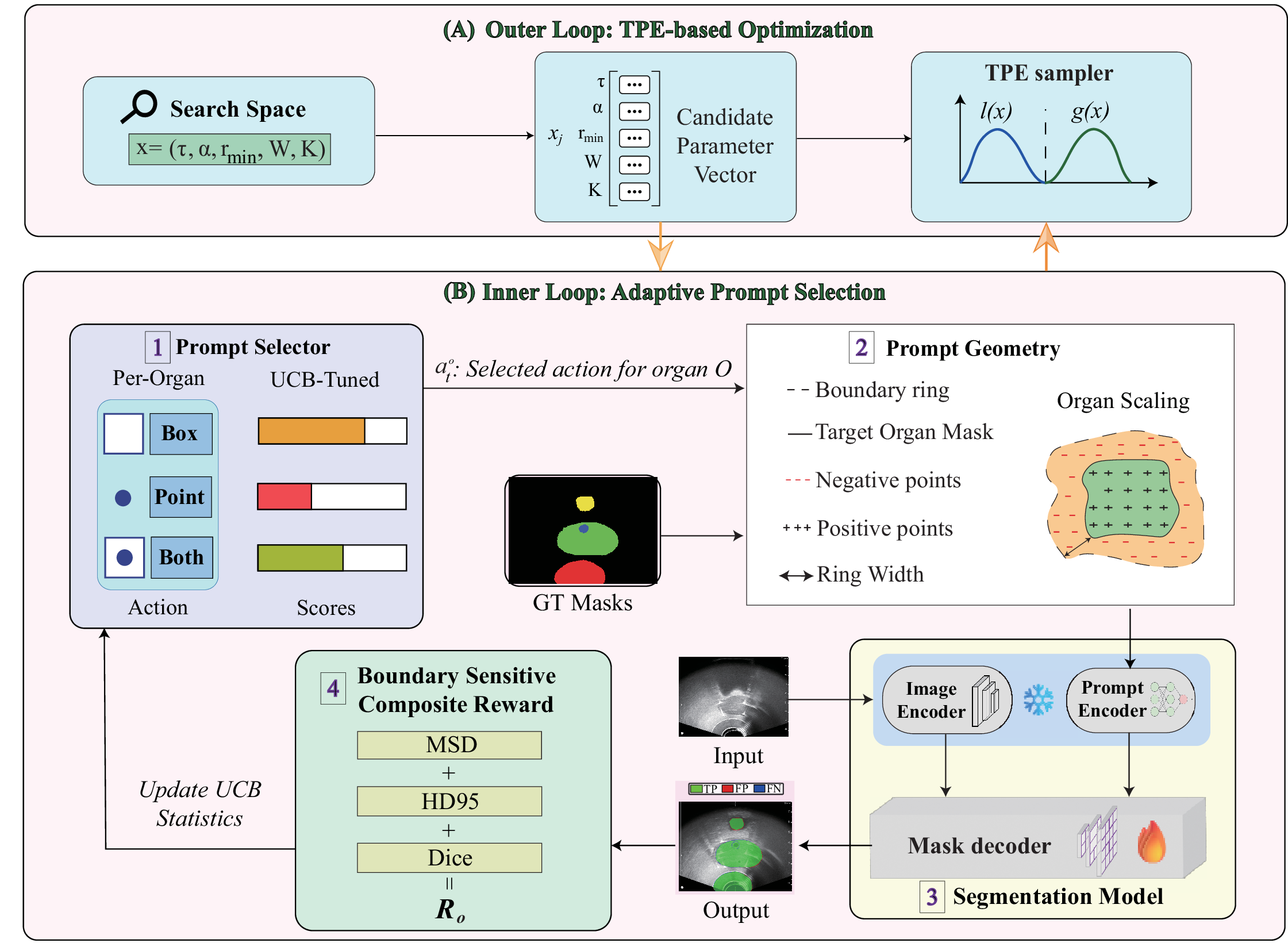}
    \caption{Overview of the BAP-MOS framework.
    (A) The outer loop runs TPE-based search over the prompt selection.
    (B) The inner loop performs UCB-based per-organ prompt selection.}
    \label{fig:bapmos_framework}
    \vspace{-0.5cm}
\end{figure}

As illustrated in Fig.~\ref{fig:bapmos_framework}, BAP-MOS is formulated as a closed-loop prompt-adaptation framework with two coupled optimization loops.  The outer loop employs Tree-structured Parzen Estimators (TPE) to optimize the prompt-selection parameter vector, while the inner loop evaluates each candidate through decoder training with UCB-based adaptive prompt selection. For each candidate vector, the resulting validation objective is fed back to the TPE sampler to iteratively refine subsequent parameter proposals until convergence. Parameters that are searched in the Loop A are listed here:




\vspace{-3pt}
\begin{equation}
\small
\mathbf{x}=(\tau,\alpha,r_{\min},W,K).
\label{eq:prompt_search_vector}
\end{equation}
\vspace{-15pt}


These parameters govern the critical mathematical interfaces between the spatial prompt generation, the non-stationary learning environment, and the UCB policy. The baseline magnitude of physical boundary errors (MSD and HD95) fluctuates drastically depending on the intrinsic acoustic shadowing and contrast of the target clinical cohort \cite{noble2006ultrasound, taha2015metrics}. Therefore, searching for an optimal value of $\tau$, which controls the reward clipping scale to satisfy the bounded-reward constraints of UCB policies \cite{auer2002finite}, is essential. The parameters $\alpha$ and $r_{\min}$ provide spatial context; these parameters define the organ-scaled negative-ring geometry. The optimal physical distance for background suppression relies heavily on domain-specific speckle noise characteristics and anatomical scales, making the automated search of these variables paramount \cite{liu2025plug}. Furthermore, the mask decoder’s convergence rate varies across different ultrasound domains. Thus, there is a need to stabilize the reinforcement learning timelines. The parameter $W$ specifies the sliding reward-memory length to track non-stationary distributions to track non-stationary distributions \cite{garivier2011upper}, and $K$ determines the number of training batches for which a selected prompt action is fixed to ensure reliable credit assignment \cite{perchet2016batched}. These parameters must be optimized to perfectly balance rapid prompt exploration against statistically stable reward estimation.\color{black}

The search ranges were determined through preliminary pilot experiments that evaluated the influence of each parameter on prompt construction, reward sensitivity, boundary accuracy, and training stability. Representative values were evaluated independently to identify stable and practically meaningful search intervals. Based on these observations, the TPE search space was defined as \(\tau\in[5,30]\), \(\alpha\in[0.05,0.15]\), \(r_{\min}\in[2,20]\), and \(W,K\in\{20,25,\ldots,100\}\). Since \(\tau\) clips the MSD component of the reward, it has the same physical unit as MSD, i.e., millimeters when pixel spacing is available and pixel-equivalent units otherwise.


For each candidate vector \(x_j\), Loop~B is executed using a fixed candidate-evaluation protocol and returns validation metrics from the corresponding adaptive prompt-selection run. Since TPE requires a scalar objective to rank candidate vectors, we define the outer-loop objective using fold-balanced validation MSD on a pre-specified primary target organ:
\begin{equation}
\small
f(x_j)
=
\frac{1}{k}
\sum_{i=1}^{k}
\overline{\mathrm{MSD}}_{o^\star,i}(x_j),
\label{eq:tpe_objective}
\end{equation}
where \(o^\star\) denotes the primary target organ for the benchmark, and \(\overline{\mathrm{MSD}}_{o^\star,i}(x_j)\) is the mean validation MSD for that organ on fold \(i\). 
The use of MSD in Eq.~\ref{eq:tpe_objective} should be interpreted as a boundary-focused scalarization for the outer TPE search, not as an exclusion of overlap or large-error criteria. Loop~A requires a single validation objective to rank candidate prompt-selection parameter vectors, and MSD provides a direct measure of average boundary disagreement, which matches the boundary-adaptive objective of BAP-MOS. At the same time, candidate evaluation inside Loop~B remains multi-metric: the prompt selector is updated using the bounded Dice--MSD--HD95 reward in Eq.~\ref{eq:composite_reward}, where Dice contributes overlap fidelity, MSD contributes average surface agreement, and HD95 contributes sensitivity to larger boundary deviations. Therefore, boundary quality is emphasized in the outer ranking objective while overlap and tail-boundary behavior remain explicitly incorporated during prompt adaptation and final evaluation.


\subsection{ Inner-Loop Adaptive Prompt Selection}

Given the candidate vector \(x_j=(\tau,\alpha,r_{\min},W,K)\), BAP-MOS performs a closed-loop training procedure consisting of four components.

\noindent\textbf{\ding{192} Prompt Selector:}
BAP-MOS maintains an independent UCB-Tuned prompt selector for each foreground organ \(o\in\mathcal{O}\). For each organ, the selector chooses an action from the discrete prompt-action space \(\mathcal{A}=\{\texttt{box},\texttt{point},\texttt{both}\}\). This per-organ design is motivated by anatomical heterogeneity in multi-organ ultrasound imaging. Larger or better-contrasted structures can often be localized effectively by global box prompts, whereas smaller or weakly visible structures may benefit from sparse point-based cues that reduce boundary ambiguity and inter-organ interference. A single shared selector would pool reward statistics across anatomically different organs and could obscure organ-specific prompt preferences. Therefore, BAP-MOS maintains independent prompt selectors so that each organ can adapt its prompt preference separately.

Rather than selecting a new prompt action for every training batch, BAP-MOS operates at decision-block granularity. At the beginning of each block, the selector assigns one action per organ, and the selected action is held fixed for \(K\) training batches. This block-wise commitment reduces prompt oscillation, provides sufficient exposure before reward assignment under evolving decoder parameters, and amortizes the cost of validation-probe evaluation.


Once a decision block is completed, the block-level reward defined in the reward-update component is committed to the selected organ-action pair and used to update the prompt selector for subsequent decisions. To balance exploration and exploitation, BAP-MOS adopts the UCB-Tuned criterion, which estimates the utility of each candidate action from its recent reward statistics. For a fixed organ, let \(Q_a(t)\) denote the empirical mean reward of action \(a\), \(N_a(t)\) denote the number of committed pulls for action \(a\), and \(t\) denote the number of committed prompt-selector updates for that organ. The variance \(\sigma_a^2(t)\) is computed from the action's sliding reward window of length \(W\) and is set to zero when fewer than two rewards are available. After deterministic warmup and minimum-pull guardrails, selection uses the UCB-Tuned score:
\begin{equation}
\small
\mathrm{UCB}_a(t)
=
Q_a(t)
+
\sqrt{
\frac{\ln(\max(1,t))}{N_a(t)}
\cdot
\tilde{V}_a(t)
}
\label{eq:ucb_score}
\end{equation}

Here, \(Q_a(t)\) favors actions with high observed reward, while the second term encourages exploration of actions with fewer committed pulls or higher uncertainty. The variance-aware term is defined as
\begin{equation}
\small
\tilde{V}_a(t)
=
\min\left(
0.25,\,
\sigma_a^2(t)
+
\sqrt{\frac{2\ln(\max(1,t))}{N_a(t)}}
\right)
\label{eq:ucb_variance}
\end{equation}
The index \(t\) counts committed selector updates for the corresponding organ and is updated only after a valid block-level reward is assigned; it is not the training-iteration index. The constant \(0.25\) follows the standard UCB-Tuned variance cap for rewards bounded in \([0,1]\), since the maximum variance of a bounded scalar reward in this range is \(1/4\).






\noindent\textbf{\ding{193}  Prompt Geometry:}
To ensure that different prompt actions are instantiated consistently across organs of varying sizes, each selected action is converted into a geometry-aware prompt representation before being passed to the frozen prompt encoder.


The box action uses the organ bounding box, with small random edge expansion during training as prompt-level augmentation. The both action combines box and point prompts into a single prompt sequence. Ground-truth masks are used only to construct prompt coordinates; they are not passed as dense mask prompts to the encoder.
The point action uses one positive point sampled from the organ foreground and three negative points sampled from a local exterior ring. 






For point-based prompting, negative prompts are drawn from an organ-scaled exterior ring rather than from arbitrary background. Let \(M_o\) denote the binary mask of organ \(o\) and \(A_o=\sum M_o\) its area in pixels. The organ-scaled ring radius is
\begin{equation}
\small
r_o
=
\max\left(
r_{\min},
\left\lfloor \alpha\sqrt{A_o}\right\rfloor
\right),
\label{eq:ring_radius}
\end{equation}
where \(\alpha\) and \(r_{\min}\) are the ring-geometry parameters in Eq.~\ref{eq:prompt_search_vector}. Scaling by \(\sqrt{A_o}\) converts the two-dimensional organ area into a one-dimensional length, so the exclusion band grows with organ size instead of using a fixed radius~\cite{lindeberg1998feature}. The negative sampling region is then obtained by dilating \(M_o\) with \(r_o\):
\begin{equation}
\small
\Omega_o
=
\operatorname{dilate}(M_o,\mathcal{S}_{r_o})
\setminus
M_o
\label{eq:ring_region}
\end{equation}

where \(\mathcal{S}_{r_o}\) is a square structuring element of size \((2r_o+1)\times(2r_o+1)\). This construction places negative prompts immediately outside the target organ, encouraging suppression of nearby background and adjacent structures.

If sparse point construction fails, for example because the organ foreground is empty or the exterior ring contains no valid background pixels, the executed prompt is downgraded to a box prompt. 


\noindent\textbf{\ding{194} Segmentation Model:}
BAP-MOS is independent of the underlying promptable foundation segmentation model and can be instantiated with SAM, MedSAM, or other compatible architectures. In this work, we adopt SAM ViT-B as the default implementation.

Following the standard fine-tuning protocol, the image encoder and prompt encoder remain frozen throughout training, while only the mask decoder is updated. The image embedding is computed once for each input image and reused across all target organs. For each organ, the effective prompt generated by the prompt-selection module is encoded by the frozen prompt encoder and decoded into an organ-specific foreground prediction. Organ-level predictions are stacked into a multi-channel output and optimized using the boundary-aware loss function:








{\small
\begin{equation}
\mathcal{L}_{\mathrm{total}}
=
\alpha(t)
\left[
\mathcal{L}_{\mathrm{CE}}(\hat{Y},Y)
+
\frac{1}{O}
\sum_{o=1}^{O}
\mathcal{L}_{\mathrm{Dice}}(\hat{Y}_o,Y_o)
\right]
+
\bigl(1-\alpha(t)\bigr)
\mathcal{L}_{\mathrm{BL}},
\label{eq:total_loss_boundary}
\end{equation}
}

where \(O\) is the number of foreground organs and \(\alpha(t)\) schedules the relative contribution of the regional and boundary terms during training. The CE--Dice component provides stable pixel-wise and region-level supervision, while \(\mathcal{L}_{\mathrm{BL}}\) introduces boundary-sensitive gradient information through ground-truth signed distance maps.






\noindent\textbf{\ding{195} Boundary-Sensitive Composite Reward.}
After every \(K\) training batches, BAP-MOS evaluates a fixed validation-probe subset using the prompt actions that were active during the completed block. For each organ, the probe computes \(\mathrm{Dice}_o\), \(\mathrm{MSD}_o\), and \(\mathrm{HD95}_o\). Valid measurements are converted into a bounded Dice--MSD--HD95 composite reward:

\begin{equation}
\small{
\begin{aligned}
R_o
=
1
-
\Bigg[
&0.3(1-\mathrm{Dice}_o)
+
0.5
\frac{\min(\mathrm{MSD}_o,\tau)}{\tau}
\\
&+
0.2
\frac{\min(\mathrm{HD95}_o,3\tau)}{3\tau}
\Bigg]
\end{aligned}
\label{eq:composite_reward}
}
\end{equation}


The reward \(R_o\) is appended only to the sliding reward window of the action selected for organ \(o\) during the completed block. The corresponding \(Q_a(t)\), \(N_a(t)\), and \(t\) values in Eq.\ref{eq:ucb_score}  are updated, and subsequent prompt selections use the revised UCB-Tuned statistics. 

\color{black}

\section{Experiments}

\subsection{Experimental Settings}
\noindent\textit{1) Datasets.}
We evaluated BAP-MOS on two ultrasound settings: an internal prostate-region TRUS benchmark and the external PFUS1 pelvic-floor corpus~\cite{solis2025pfus1}. The internal benchmark comprises one phantom simulation (91 foreground slices) and two clinical cohorts (case~1: 76; case~2: 78 slices), reflecting a prostate radiotherapy setting in which the target and surrounding organs at risk share a common label space of Bladder, Prostate/PTV, Rectum, and Urethra. Slices were converted to 8-bit multiclass masks, with simulation images resampled to the clinical spacing of 0.159072\,mm/pixel. Each cohort was split 70/15/15 (organ-presence stratification, seed~42); training and validation slices were pooled across cohorts (168 and 38), while the three test sets were kept separate (14 simulation, 12 and 13 clinical). PFUS1 provides 14{,}852 midsagittal frames from 110 patients with eight annotated structures; the same 70/15/15 split at the patient level (acquisition-length stratification) yielded 77/17/16 patients (10{,}576/2{,}175/2{,}101 frames). Its distance metrics use unit spacing and are reported in pixel-equivalent units.

\noindent\textit{2)Training settings.}
All experiments followed the decoder-only fine-tuning protocol in Section~\ref{sec:method}, using Adam (learning rate \(10^{-4}\)) and evaluated with Dice, MSD, and HD95. Optuna TPE parameter search ran 100 trials (seed~42): the first 20 were fixed startup configurations (one baseline plus 19 Latin-hypercube samples), and the rest were adaptively proposed from completed validation objectives. Pooled TRUS runs used up to 300 epochs (early-stopping patience~40) and PFUS1 bladder runs up to 100 epochs (patience~20). Experiments ran on heterogeneous NVIDIA GPUs (H100, A100, A40); GPU type affected only throughput, not results, as the protocol, optimizer, seed, validation, early-stopping, and metrics were fixed across platforms.

\begin{table}[t]
\centering
\scriptsize
\begin{threeparttable}
\caption{Quantitative results on the prostate-region TRUS benchmark; the bottom two rows (\(\dagger\)) report external PFUS1 generalization.}
\label{tab:external_baselines_prostate}
\setlength{\tabcolsep}{5pt}
\begin{tabular}{lccc}
\toprule
\textbf{Method} & \textbf{Dice\,$\uparrow$} & \textbf{HD95\,$\downarrow$} & \textbf{MSD\,$\downarrow$} \\
\midrule
nnU-Net
& 0.962 {\tiny$\pm$0.002}
& 1.113 {\tiny$\pm$0.131}
& 0.432 {\tiny$\pm$0.035} \\
U-Net
& 0.965 {\tiny$\pm$0.004}
& 0.932 {\tiny$\pm$0.158}
& 0.369 {\tiny$\pm$0.045} \\

SAM
& 0.981 {\tiny$\pm$0.002}
& 0.527 {\tiny$\pm$0.062}
& 0.221 {\tiny$\pm$0.001} \\

MedSAM
& 0.941 {\tiny$\pm$0.002}
& 2.736 {\tiny$\pm$0.725}
& 0.868 {\tiny$\pm$0.100} \\

\textbf{BAP-MOS (SAM)}
& \textbf{0.982 {\tiny$\pm$0.001}}
& \textbf{0.482 {\tiny$\pm$0.016}}
& \textbf{0.204 {\tiny$\pm$0.023}} \\

\textbf{BAP-MOS (MedSAM)}
& \textbf{0.979 {\tiny$\pm$0.006}}
& \textbf{0.577 {\tiny$\pm$0.032}}
& \textbf{0.229 {\tiny$\pm$0.013}} \\
\hline\hline
FPN~\cite{garcia2024applicability}$^{\dagger}$
& 0.710
& --
& -- \\
\textbf{BAP-MOS (MedSAM)$^{\dagger}$}
& \textbf{0.849 {\tiny$\pm$0.007}}
& \textbf{10.062 {\tiny$\pm$0.55}}
& \textbf{5.034 {\tiny$\pm$0.015}} \\
\bottomrule
\end{tabular}
\begin{tablenotes}
\footnotesize
\item[$\bullet$] TRUS distance metrics are reported in millimeters.
\item[$\dagger$] External PFUS1 pelvic-floor dataset, bladder; distances in pixel-equivalent units. FPN results from~\cite{garcia2024applicability} report Dice only.
\end{tablenotes}
\end{threeparttable}
\vspace{-0.5cm}
\end{table}


\subsection{Quantitative Results }

On the pooled prostate-region TRUS benchmark (Table \ref{tab:external_baselines_prostate}),
BAP-MOS(SAM) attains the best score on every metric. First, it comprehensively surpasses the strongest conventional baseline U-Net (Dice from 0.965 to 0.982, HD95 from 0.932 to 0.482, MSD from 0.369 to 0.204). Second, it markedly improves the
boundary-localization accuracy of SAM: although BAP-MOS (SAM) shows almost no gain in Dice (\(0.981\) vs.\ \(0.982\), since SAM's overlap is already near saturation and further improvement is
difficult), it achieves clear gains on the other two metrics, reducing HD95 by \(8.5\%\) (from 0.527 to 0.482) and MSD by \(7.7\%\) (from 0.221 to 0.204). This shows that BAP-MOS can substantially enhance a model's boundary-localization ability while preserving its high overlap. In contrast, MedSAM was originally even weaker than the conventional U-Net baseline, yet after applying BAP-MOS its performance rises to near the SAM level, with HD95 and MSD improving several-fold (from 2.736 to 0.577 and from 0.868 to 0.229). This weakness stems from a domain mismatch in MedSAM's pretraining. Although MedSAM includes ultrasound in its training distribution, ultrasound accounts for only 6.01\% of its training image--mask pairs, a dataset dominated by CT and MRI, and its ultrasound tasks are limited to single-target, standard-view acquisitions such as breast and thyroid nodules, fetal head, kidney, and cardiac chambers~\cite{ma2024segment}. None involves the low-contrast, speckle-heavy, multi-organ transrectal setting considered here. Consequently, MedSAM's medical-imaging priors transfer poorly to multi-organ TRUS, yielding weaker boundary delineation than SAM, whose broader natural-image pretraining is less tied to these structured medical scenarios. That adaptive per-organ prompting nonetheless recovers most of this gap suggests the limitation lies in the fixed domain prior rather than in the promptable framework itself.

To assess generalization beyond the internal prostate-region TRUS domain, we further evaluated BAP-MOS (MedSAM) on PFUS1, reported in the last row of Table~\ref{tab:external_baselines_prostate}. PFUS1 is a considerably harder benchmark, jointly annotating eight densely packed pelvic-floor structures under low tissue contrast and dynamic deformation~\cite{solis2025pfus1}. For this external validation we adopt only the MedSAM backbone, since MedSAM is purpose-built for medical image segmentation, so the further-dataset evaluation focuses on adapting this medical foundation model. BAP-MOS (MedSAM) attains a bladder Dice of \(0.849\) on this setting, well above the \(0.71\) reported for FPN, the best-performing model in a prior CNN benchmark on this data~\cite{garcia2024applicability}. In addition, this is the first promptable foundation model evaluated on this benchmark, confirming that the overlap-level advantage of BAP-MOS transfers to an external medical ultrasound domain.


\subsection{Visual Results}

\begin{figure}[t]
    \includegraphics[width=0.48\textwidth]{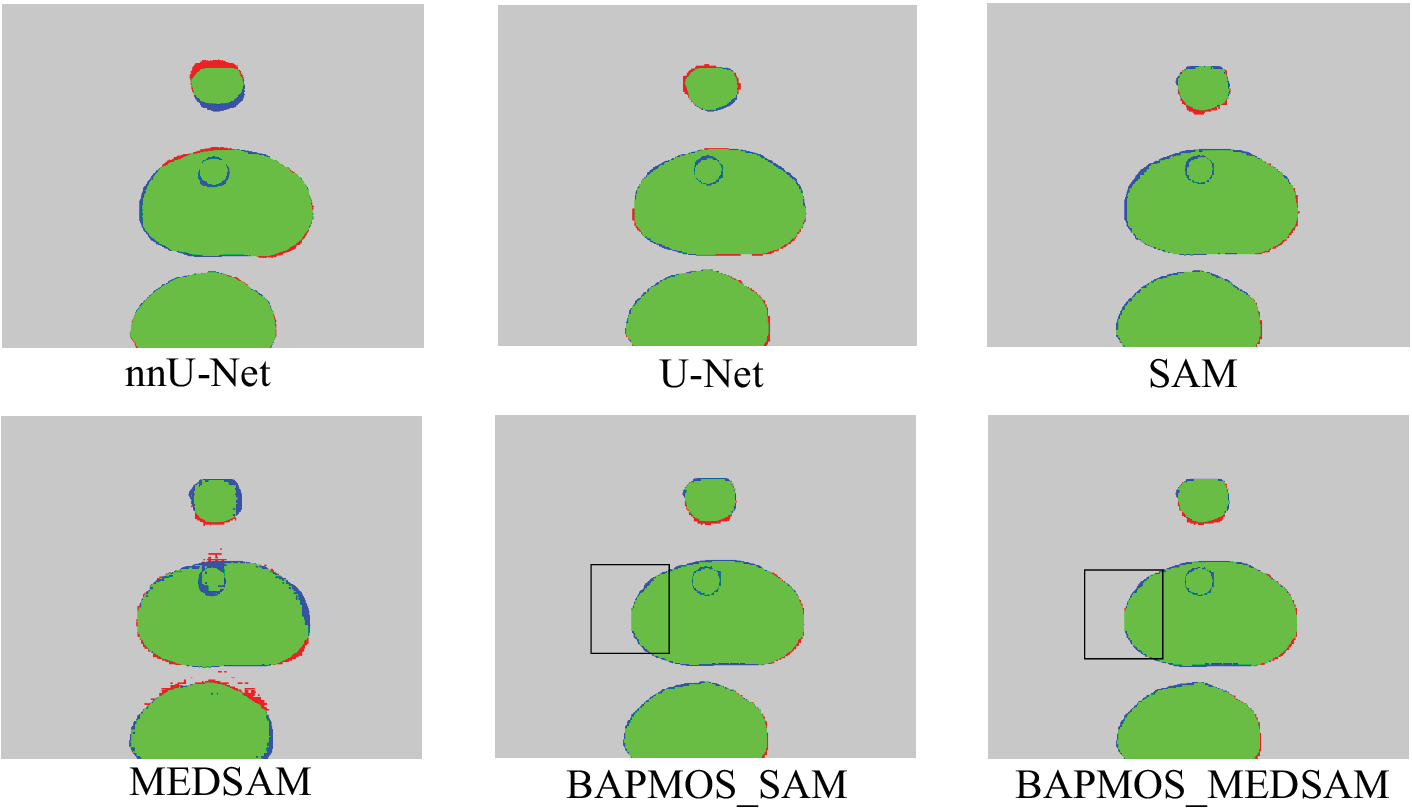}
    \caption{Boundary deviation of BAP-MOS and baseline methods.}
    \label{fig:qualitative_trus}
    \vspace{-0.5cm}
\end{figure}


Figure~\ref{fig:qualitative_trus} presents a qualitative comparison of BAP-MOS and baseline methods on a representative slice. For each method, the corresponding TP/FP/FN error map is shown. Compared with fixed-prompt SAM/MedSAM and fully supervised baselines, BAP-MOS produces anatomically localized masks with reduced boundary disagreement around the target structures. The error maps show that remaining false-positive and false-negative regions are concentrated mainly near organ interfaces and smaller adjacent structures, rather than reflecting large-scale organ mislocalization. This visual behavior is consistent with the quantitative results, where MSD and HD95 complement Dice by capturing residual boundary-level errors.


\subsection{Ablation Studies}

\subsubsection{Search map for the TPE}
We first analyzed the parameter search based on TPE to examine how the selected suggestion vector is obtained. The first 20 trials correspond to the fixed startup catalog (one baseline setting and 19 Latin-hypercube samples), while the remaining trials are adaptive TPE proposals generated from completed validation objectives. Rather than enumerating the search space uniformly, after the startup phase TPE revisits and refines parameter regions associated with lower validation objective values, and the selected final configuration was obtained at trial~61, with \((\tau,\alpha,r_{\min},W,K)=(26,0.108,10,50,70)\).

Next, to verify the advantages of TPE, we compared TPE with greedy, random, and heuristic algorithms.The selected parameter values and search costs of different search strategies are reported in Table~\ref{tab:search_strategy_comparison}. TPE achieved the lowest validation objective among the compared strategies, with a best objective of \(0.1917\,\mathrm{mm}\), compared with \(0.1940\,\mathrm{mm}\) for random search, \(0.1961\,\mathrm{mm}\) for the simple heuristic, and \(0.1964\,\mathrm{mm}\) for the greedy strategy. Although these objective differences are modest, TPE also required less total compute than random search and greedy search, reducing total compute from \(66.67\) to \(56.80\) GPU-hours relative to random search and from \(95.74\) to \(56.80\) GPU-hours relative to greedy search. These results suggest that TPE provides a reproducible and compute-bounded mechanism for selecting the prompt-selection vector. 

\begin{figure}[t]
    \includegraphics[width=0.48\textwidth]{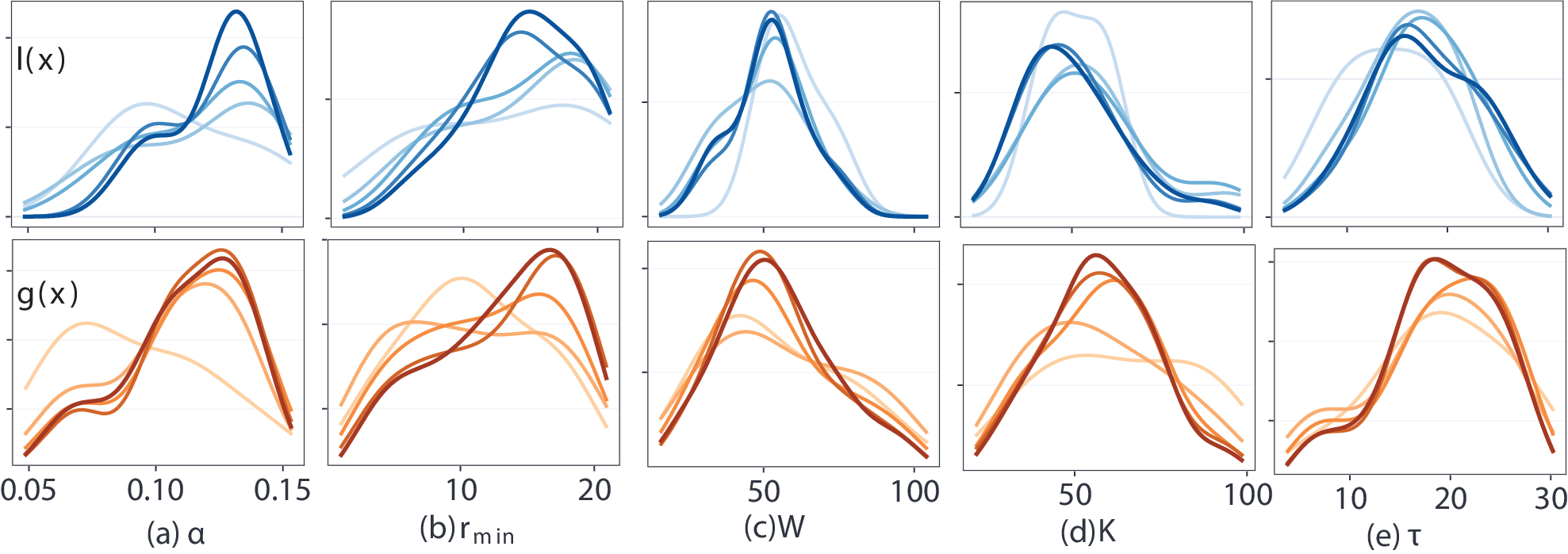}
    \caption{Evolution of the TPE probability distributions for different hyperparameters during the search process. Darker curves correspond to later optimization iterations.}
    \label{fig:tpe}
    \vspace{-0.2cm}
\end{figure}

Finally, Fig.~\ref{fig:tpe} provides the insight into how TPE arrived at this configuration. At each step TPE refits $l(x)$ and $g(x)$, so the progression from lighter to darker curves follows the stabilization of the estimated distributions. Across the hyperparameters $l(x)$ changes from early estimates towards concentrated modes as trials accumulate, indicating that the search progressively converges on influential regions of the space. The extent of the contraction varies by parameter, with some densities sharpening into well defined peaks and other remaining comparatively broad, reflecting how each parameter influences the objective. 

\begin{table}[t]
\centering
\scriptsize
\begin{threeparttable}
\caption{Prompt-selection parameter search: selected configuration and search cost across strategies.}
\label{tab:search_strategy_comparison}
\setlength{\tabcolsep}{5pt}
\begin{tabular*}{\columnwidth}{@{\extracolsep{\fill}}lccccccc@{}}
\toprule
\textbf{Strategy}
& \(\boldsymbol{\tau}\)
& \(r_{\min}\)
& \(\alpha\)
& \(W\)
& \(K\)
& \makecell{\textbf{Obj.}}
& \makecell{\textbf{Cost}} \\
\midrule
TPE~\cite{bergstra2011algorithms}
& 26 & 10 & 0.108 & 50 & 70 & \textbf{0.1917} & \textbf{56.8} \\
Random~\cite{bergstra2012random}
& 5 & 17 & 0.147 & 35 & 35 & 0.1940 & 66.7 \\
Heuristic~\cite{bergstra2011algorithms}
& 21 & 14 & 0.052 & 50 & 50 & 0.1961 & 136.7 \\
Greedy~\cite{wright2015coordinate}
& 23 & 17 & 0.057 & 80 & 25 & 0.1964 & 95.7 \\
\bottomrule
\end{tabular*}
\end{threeparttable}
\vspace{-0.3cm}
\end{table}

\subsubsection{Learning dynamics}
\begin{figure}[t]
    \centering
    \includegraphics[width=0.49\textwidth]{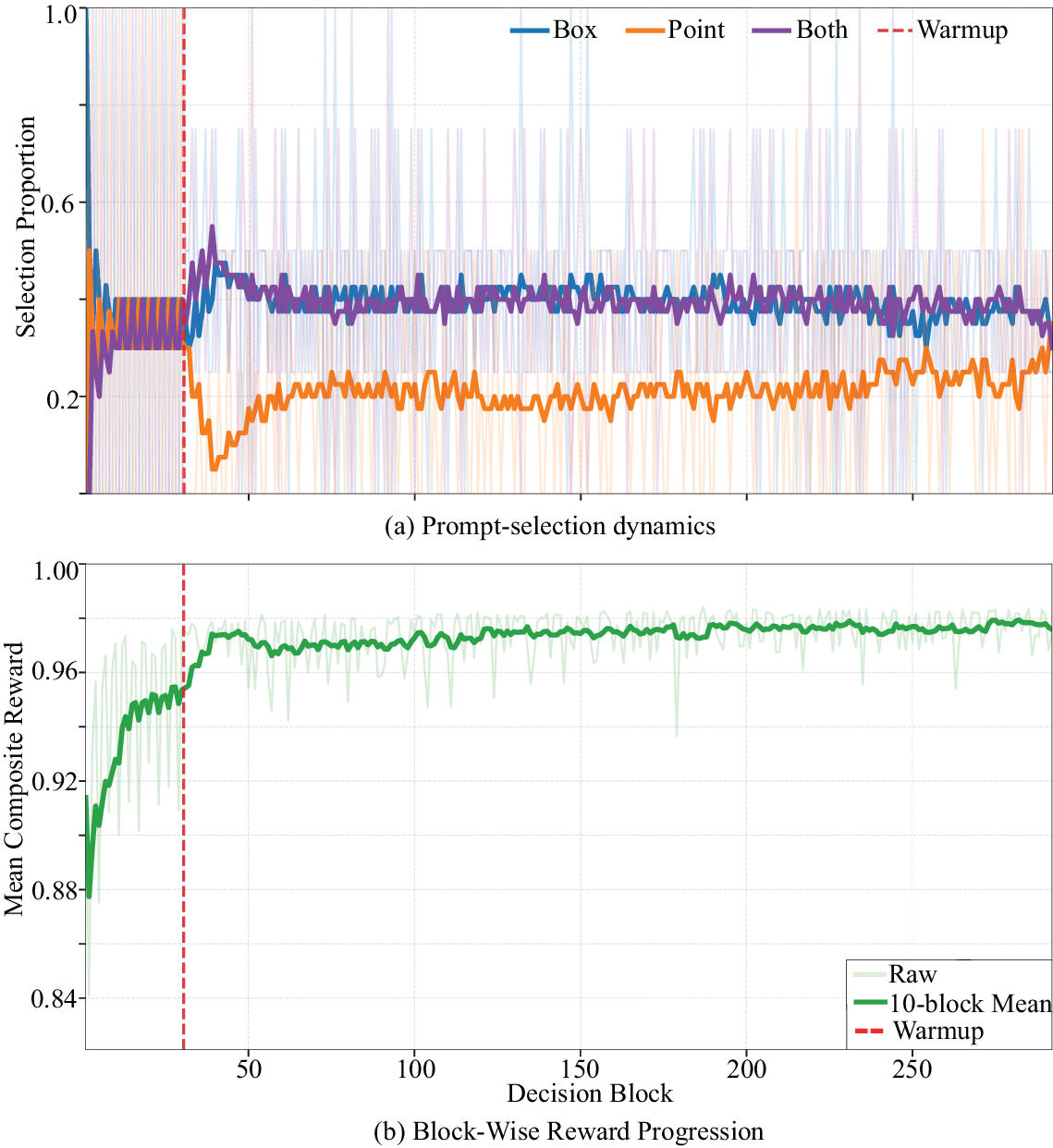}
    \caption{Prompt-selection learning dynamics for a representative pooled TRUS BAP-MOS (SAM) run. 
    (a) Running selection proportions for box, point, and combined prompts; faint background marks show raw block-level actions, and the dashed red line indicates the warmup end. 
    (b) Block-wise composite reward, shown as raw reward and a 10-block moving average.}
    \label{fig:learning_dynamics}
    \vspace{-0.5cm}
\end{figure}

Figure~\ref{fig:learning_dynamics} shows prompt-selection dynamics for a representative pooled TRUS BAP-MOS run. During warmup, the selector allocates roughly uniformly across arms, providing the initial exploratory reward estimates for the policy. Once exploitation begins, the selection share of point-only prompting drops sharply before the exploration term partially restores it, while box and combined prompts converge to a stable dominant allocation. This dip-and-recovery pattern reflects the intended exploration-exploitation balance: the policy commits to a dominant prompting mode without permanently starving weaker arms, allowing continued reassessment as segmentation feedback accumulates. The block-wise reward tracks this behavior, improving rapidly at first then flattening once the selection policy stabilizes.

\subsubsection{Bandit cost and average Q}
To verify that this behavior was consistent across runs, we analyzed the saved bandit states from the best validation checkpoints for seeds 42, 43 and 44. As summarized in Table~\ref{tab:prompt_preference_q_values}, box and combined prompting achieved nearly identical final selection rates and learned Q-values, whereas point-only prompting was selected less frequently with a correspondingly lower Q-value. This agreement between selection rate and Q-value suggests that the selector converged to a stable preference for global or combined spatial guidance. Table~\ref{tab:bandit_cost} shows the corresponding bandit update cost for each run. Each run completed roughly 200 to 250 blocks and fewer than 1000 total arms pulls with wall times between two to four hours.

\begin{table}[t]
\centering
\footnotesize
\caption{Prompt-selection controller cost and learned preferences on the pooled TRUS benchmark.}
\label{tab:controller_cost_preference}

\begin{subtable}[t]{0.33\linewidth}
\centering
\setlength{\tabcolsep}{3pt}
\begin{tabular}{lccc}
\toprule
\textbf{Seed} & \textbf{Blocks} & \textbf{Pulls} & \textbf{Time(m)} \\
\midrule
42 & 196 & 784 & 112.90 \\
43 & 247 & 988 & 256.74 \\
44 & 244 & 976 & 257.37 \\
\bottomrule
\end{tabular}
\caption{Bandit update cost.}
\label{tab:bandit_cost}
\end{subtable}
\hfill
\begin{subtable}[t]{0.62\linewidth}
\centering
\setlength{\tabcolsep}{3pt}
\begin{tabular}{lcc}
\toprule
\textbf{Arm} & \textbf{Sel.(\%)} & \textbf{\(Q\)} \\
\midrule
Box   & \(39.72{\scriptstyle\pm0.16}\) & \(0.981{\scriptstyle\pm0.001}\) \\
Point & \(20.81{\scriptstyle\pm0.51}\) & \(0.925{\scriptstyle\pm0.002}\) \\
Both  & \(39.46{\scriptstyle\pm0.36}\) & \(0.981{\scriptstyle\pm0.001}\) \\
\bottomrule
\end{tabular}
\caption{Final prompt preferences and \(Q\)-values.}
\label{tab:prompt_preference_q_values}
\end{subtable}

\vspace{-0.8cm}
\end{table}

\color{black}

\section{Conclusion}

\noindent We presented \textbf{BAP-MOS}, a boundary-adaptive prompting framework for multi-organ ultrasound segmentation. Instead of using a fixed prompt type throughout training, BAP-MOS formulates prompt selection as an organ-specific bandit problem and learns whether box, point, or combined prompts are most effective for each anatomical structure. The framework couples TPE-based parameter search with per-organ UCB-Tuned prompt optimization, keeping the SAM image and prompt encoders frozen and updating only the mask decoder. On the TRUS benchmark, BAP-MOS achieved the strongest performance across Dice, MSD, and HD95, with especially large gains in boundary-distance metrics, and on external PFUS1 it further improved bladder-region overlap under MedSAM initialization. These results show that prompt allocation is an important design variable for promptable ultrasound segmentation: by adapting prompt type during decoder fine-tuning, BAP-MOS improves boundary-sensitive segmentation without modifying the foundation-model backbone.

\balance
\tiny
\bibliographystyle{IEEEtran}
\bibliography{references}


\end{document}